\documentclass{article}

\usepackage[final]{neurips_2026}

\makeatletter

\providecommand{\@trackname}{}

\makeatother
\usepackage[utf8]{inputenc}
\usepackage[T1]{fontenc}
\usepackage{hyperref}
\usepackage{url}
\usepackage{booktabs}
\usepackage{amsfonts}
\usepackage{amsmath}
\usepackage{amssymb}
\usepackage{mathtools}
\usepackage{amsthm}
\usepackage{microtype}
\usepackage{subcaption}
\usepackage{graphicx}
\usepackage{xcolor}
\usepackage[table]{xcolor}
\usepackage[capitalize,noabbrev]{cleveref}
\usepackage{float}
\floatstyle{plaintop}
\restylefloat{table}
\usepackage[tableposition=top]{caption}

\theoremstyle{plain}

\theoremstyle{definition}

\theoremstyle{remark}

\usepackage[textsize=tiny]{todonotes}

\newcommand{\pnplaffil}{%
  PNPL\,\raisebox{-0.15ex}{\includegraphics[height=2.2ex]{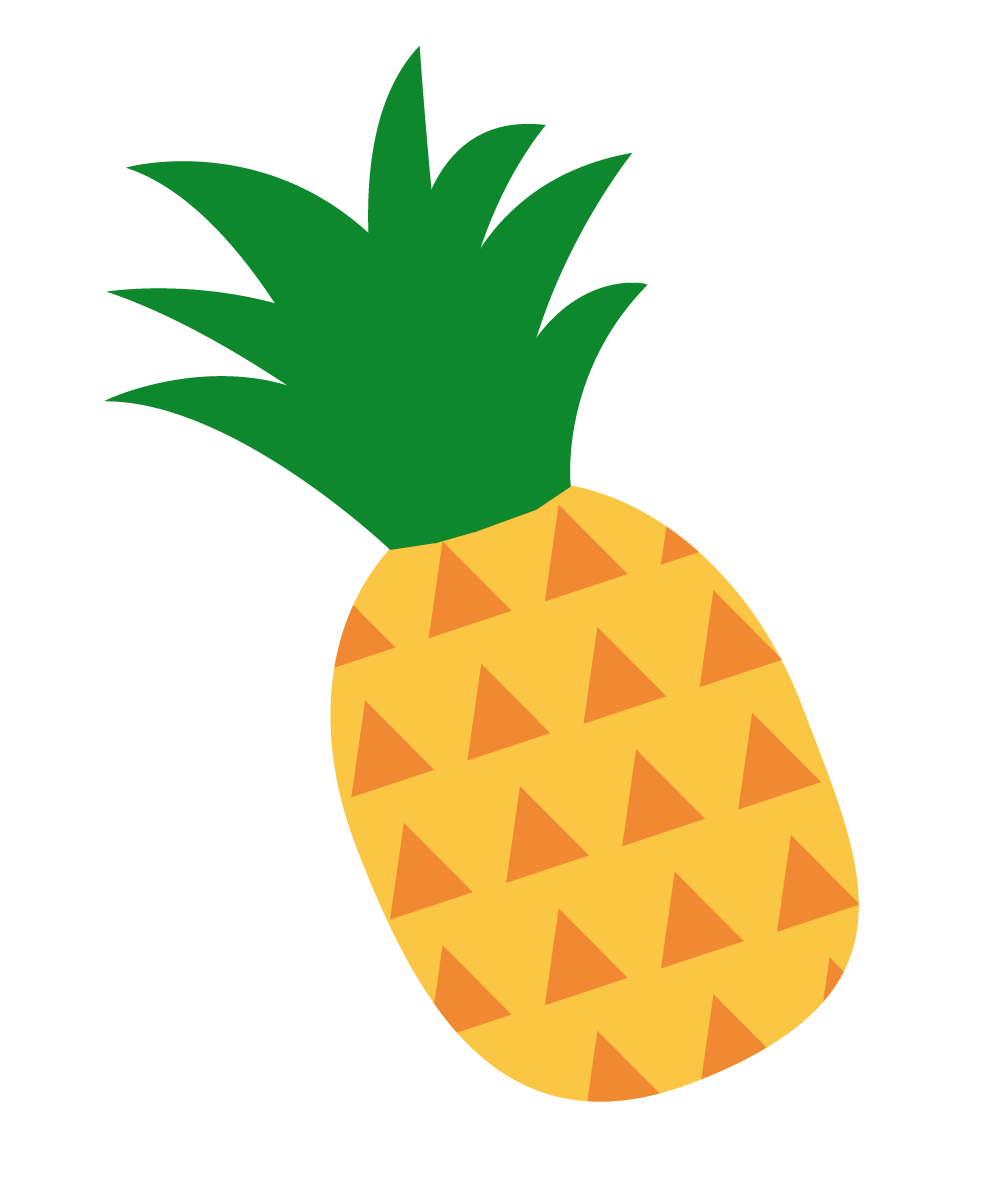}} University of Oxford%
}

\title{The Semantic Bottleneck: Leveraging Semantic Representations for Non-Invasive Speech Decoding}

\author{
  Gilad D. Landau \\
  \pnplaffil\\
  \texttt{gilad.landau@jesus.ox.ac.uk}
  \And
  Dulhan Jayalath \\
  \pnplaffil
  \And
  Oiwi Parker Jones \\
  \pnplaffil
}
\begin{document}

\maketitle
\begin{abstract}
Non-invasive speech decoding remains constrained by the low signal-to-noise ratio of neural recordings, which makes fine-grained reconstruction of phonemes or individual words difficult. Motivated by neuroscientific evidence that high-level semantic representations are distributed across cortical regions and evolve over slower temporal scales, we hypothesize that semantic content may provide a more suitable target for non-invasive decoding than low-level acoustic or lexical features. We introduce \textbf{Brain2Semantics2Text}, a method that reconstructs text through an intermediate semantic embedding space. Our model maps sentence-level MEG responses into a semantic manifold and then inverts the predicted embeddings into natural language. This semantic bottleneck enables recovery of high-level meaning without word-level alignment. We describe the core principles of the approach, its implementation, and the strategies used to mitigate the challenges of learning a reliable neural-to-semantic mapping. Finally, we compare against prior non-invasive \textit{Brain2Text} methods and show improved sentence-level results.
\end{abstract}

\section{Introduction}

Speech decoding BCIs have long been a sought-after goal in both neuroscience and healthcare. Recent progress in \textit{invasive} \textit{Brain2Text} systems has demonstrated the feasibility of translating neural activity into language \citep{Anumanchipalli2019_SpeechSynthesis,Moses2021_Neuroprosthesis,Willett2023_HighPerf,Card2024_Neuroprosthesis}. However, these invasive approaches require surgical implantation of intracranial electrodes, creating a strong incentive to develop non-invasive speech decoding solutions that are safer, more accessible, and easier to deploy.

\looseness=-1 Extending this paradigm to non-invasive systems remains a major challenge, largely due to their inherently lower signal-to-noise ratios. Lower signal fidelity makes it difficult to reliably decode fine-grained linguistic units such as phonemes or individual words. In contrast, higher-order contextual semantic representations are spatially distributed across the cortex \citep{huth2016natural_semantic_maps}, exhibit substantial redundancy, and evolve over slower temporal scales \citep{Gwilliams2025_HierarchicalDynamicCoding}. These properties may make semantic representations particularly amenable to decoding from non-invasive neural recordings, whose spatial and temporal characteristics are better matched to distributed, slowly evolving signals.

\begin{figure}[t]
    \centering
    \includegraphics[
        width=\textwidth,
        trim=20 0 20 0,
        clip
    ]{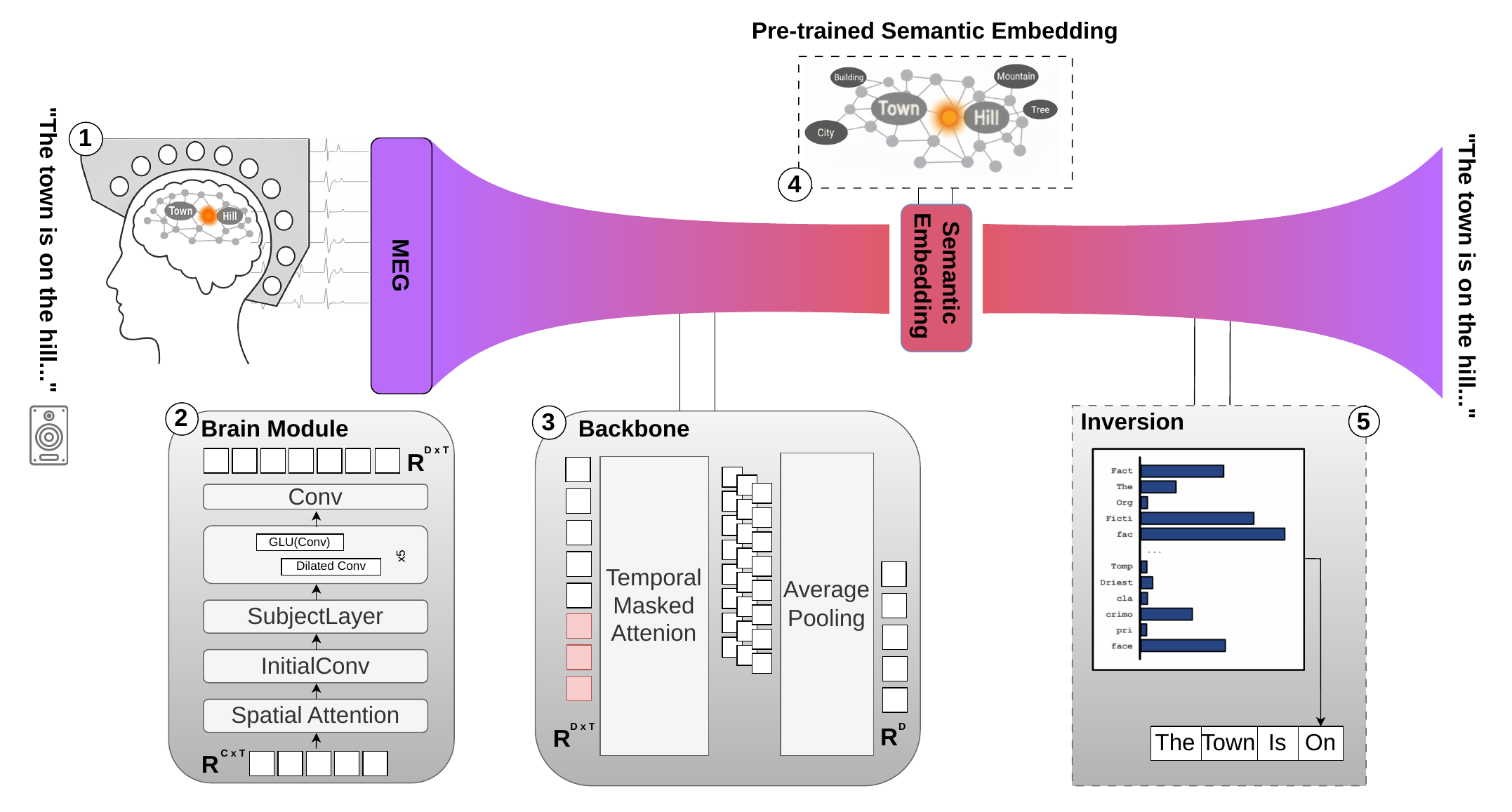}
    \caption{
    \textbf{(1)} MEG responses are collected while the participant listens to spoken sentence stimuli.
    \textbf{(2)} The Brain Module extracts time-resolved neural features using spatial attention and dilated temporal convolutions.
    \textbf{(3)} A backbone with temporal masked attention pools the neural sequence into a fixed-dimensional vector.
    \textbf{(4)} The vector is trained to be aligned with a pre-trained sentence-embedding space representing semantic content.
    \textbf{(5)} The predicted semantic embedding is inverted back to text.
    }
    \label{fig:teaser}
\end{figure}

Recently, a growing body of work in both the speech decoding literature and the neuroscience of language has converged on the view that speech comprehension and production rely on a hierarchical organization of neural representations \citep{Gwilliams2025_HierarchicalDynamicCoding}. Lower levels of this hierarchy are dominated by auditory and articulatory representations closely tied to the acoustic structure of speech, while progressively higher levels abstract away from these surface properties, giving rise to representations that are less dependent on specific phonetic or lexical features and more closely associated with the meaning of speech (\citet{Gwilliams2025_ComputationalArchitecture}, \citet{Goldstein2025_TemporalStructure}).

 Within this framework, the success of invasive approaches can be largely attributed to their ability to directly access high-fidelity neural signals from the auditory and articulatory components of speech processing, which occupy the lower levels of the cortical hierarchy (often supplemented by post-hoc language models that guide generation; \citep{Willett2024_BrainToTextBenchmark}). The same hierarchical view also motivates directly targeting higher-level semantic representations as a distinct and parallel neural signal. If such representations can be reliably decoded, their defining properties—slow temporal dynamics, distributed cortical organization, and representational redundancy—are better matched to the spatial and temporal characteristics of non-invasive modalities such as fMRI, MEG, and EEG.

This work targets semantic representations in brain activity, focusing on higher-level stages of the speech-processing hierarchy. \textit{Brain2Semantics2Text} maps sentence-length MEG responses during heard speech into a pretrained semantic embedding space, from which text is subsequently reconstructed. 
By constraining neural decoding to pass through this \textit{semantic bottleneck}, the method aims to shift the objective toward higher-level speech representations that carry information about sentence-level meaning.

Several aspects distinguish our method from previous work on \emph{Brain2Text}:
\begin{itemize}
    \item \textbf{Semantic embedding inversion.}
    We build on recent advances in semantic embedding inversion \citep{Morris2023}, which enables the reconstruction of text from semantic embeddings either as an intrinsic property of newer embedding models \citep{Duquenne2023SONAR} or via general inversion techniques applicable to arbitrary pre-trained semantic spaces \citep{Jha2025UniversalGeometry}. This allows us to frame speech decoding as semantic reconstruction rather than word or phoneme prediction.

\item \textbf{Sentence-level semantic decoding.}
Our method operates directly at the sentence level, targeting compositional semantic representations near the top of the speech-processing hierarchy. This formulation shifts the decoding problem away from exact word or phoneme recovery and toward reconstruction of the intended semantic content. As a result, it avoids dependence on precise word-level alignment and closed-vocabulary supervision, both of which are difficult to assume in realistic settings. Although full-sentence decoding from MEG is ambitious, sentence-level semantic decoding is well matched to the distributed and temporally extended nature of high-level language representations, making it a promising direction for future non-invasive communication systems.

    \item \textbf{Semantic decoding from MEG.}
    Unlike most prior semantic decoding work, which relies on the high spatial resolution of fMRI \citep{Tang2023}, we leverage the largest single-subject heard-speech MEG dataset of its kind to date. Decoding semantic-level information from MEG enables the joint exploitation of slow, distributed semantic signals and local, high-frequency neural activity within the same recordings, opening new possibilities for improved speech decoding.
\end{itemize}

\section{Related Work}

\paragraph{Semantic decoding.}
Semantic representations have previously been used as an intermediate target for non-invasive language decoding. \citet{Pereira2018UniversalDecoder} demonstrated that fMRI responses to sentences could be mapped into semantic embedding spaces, providing early evidence that distributed neural activity can be aligned with sentence-level meaning. More recently, \citet{Tang2023} reconstructed continuous perceived and imagined language from fMRI by mapping distributed cortical responses into semantic representations and using these representations to constrain language generation. These approaches exploit the high spatial resolution of fMRI to recover distributed semantic information, but sacrifice the temporal resolution available in electrophysiological recordings.

\citet{Wang2023SemanticReconstructionMEG} extended semantic reconstruction to MEG, demonstrating that semantic information can also be recovered from temporally resolved non-invasive recordings. Their approach, however, operates at the word level, reconstructing a temporally aligned sequence of contextual word embeddings that is subsequently used to generate continuous text. In contrast, our method treats the entire sentence as the unit of decoding, mapping sentence-length MEG responses directly into a single pre-trained sentence-level semantic embedding. This formulation makes the semantic representation itself the decoding bottleneck and removes the need for word-level alignment.

\paragraph{Auditory and speech-based MEG decoding.}
A complementary line of work maps MEG activity to representations closely tied to the acoustic structure of speech. \citet{Defossez2023DecodingSpeech} aligned MEG responses with Wav2Vec representations \citep{Baevski2020wav2vec2}, establishing a contrastive-learning framework and neural architecture that have influenced subsequent MEG decoding systems. More recent approaches align MEG with auditory representations \citep{Yang2024MAD}, adapt Whisper for neural speech decoding \citep{yang2024neuspeech}, or incorporate neural signals into multimodal foundation-model architectures \citep{yang2024neugpt}. These approaches exploit neural information associated with the acoustic realization of speech, whereas our method deliberately targets higher-level semantic content.

Within this line of work, BrainECHO \citep{Li2024BrainECHO} provides the closest comparison to our method. Like \textit{Brain2Semantics2Text}, BrainECHO operates at the sentence level and does not require word-level alignment, but the two methods differ in the representation through which decoding proceeds. BrainECHO maps neural activity into a vector-quantized audio-spectrogram latent space before generating text with Whisper, whereas our method maps MEG directly into a sentence-level semantic embedding. BrainECHO therefore provides a particularly informative baseline for evaluating the use of semantic, rather than acoustic, representations as a bottleneck for sentence-level decoding.

\paragraph{Word-level classification.}
Semantic representations have also been used for word-level MEG decoding. \citet{dAscoli2024DecodingWords} achieve strong closed-vocabulary word classification by aligning MEG responses with lexical semantic embeddings augmented by sentence context. Their approach demonstrates the utility of semantic representations for MEG decoding, but relies on exact word-level timing and formulates decoding as classification among candidate words. Our approach instead targets a single compositional representation of the complete sentence, removing the requirement for word-level alignment at the cost of a substantially less constrained reconstruction problem.

\section{Method}
The \textit{Brain2Semantics2Text} method operates in two stages. In the first stage, MEG neural responses corresponding to continuously presented spoken sentences are mapped to vector representations in the pre-trained semantic embedding space. Training is guided by objectives that encourage alignment with the target embeddings while preserving their global statistical structure. In the second stage, the predicted semantic embedding is inverted into natural language using a pre-trained inversion model~\citep{Morris2023} that reconstructs text from semantic vectors.

\subsection{Backbone}
\paragraph{MEG-to-semantic mapping.}
The MEG input signal $\mathbf{x} \in \mathbb{R}^{C \times T}$, where $C$ denotes the number of sensors and $T$ the number of temporal samples, is first processed by a spatial attention module, followed by an initial $1 \times 1$ convolution that projects the sensor dimension into a latent feature space. The resulting representation is then passed through a stack of dilated temporal convolutional blocks. Each block consists of dilated convolutions equipped with residual connections and gated linear units (GLUs), enabling the model to capture long-range temporal dependencies.

A subject-specific layer can optionally be inserted after the initial projection to model inter-subject variability. In the experiments reported here all data originate from a single subject.
\paragraph{Temporal aggregation.}
To obtain a fixed-dimensional semantic representation from the time-resolved features, we use a 4‑head self‑attention Transformer over the temporal axis.
We then pool over time with a masked mean
\[
\hat{\mathbf{y}} = \frac{\sum_{t=1}^{T} m_t\, H_t}{\sum_{t=1}^{T} m_t} \in \mathbb{R}^{d},
\]
where $m_t$ is a temporal mask based on the true segment lengths, preventing length‑related surface confounders from influencing the pooled embedding.

\subsection{Semantic Embeddings}

Many semantic embedding models are available, with different architectures, training objectives, and benchmark performance \citep{muennighoff2022mteb}. 
For \textit{Brain2Semantics2Text}, however, standard evaluations on semantic similarity or retrieval tasks provide only partial guidance. 
Our method requires the embedding space to function as an invertible bottleneck between MEG responses and text, which introduces additional constraints beyond general semantic performance. 
In particular, the embedding must have an available inversion mechanism and must support reliable reconstruction from both exact text embeddings and imperfect neural predictions. 
We therefore evaluate candidate embedding spaces according to four criteria.
\paragraph{Expressivity.}
The embedding space should preserve enough information about the original sentence to support reconstruction.
We measure this using a round-trip reconstruction test, in which each sentence is embedded and then inverted back into text.
Higher reconstruction quality indicates that more sentence-level information is retained by the embedding and its inversion procedure.
\paragraph{Soft reversibility.}
\label{par:soft_reversibility}
At test time, the vectors being inverted are not exact text embeddings, but embeddings predicted from noisy MEG responses.
The embedding space should therefore be robust to prediction error: vectors near the target should still invert to text with similar meaning.
We assess this by perturbing target embeddings and calculating the relation between the introduced noise and reconstruction fidelity.
\paragraph{Length bias.}
The embedding should encode sentence meaning without being dominated by surface-level properties such as sentence length.
This is particularly important because stimulus duration and sentence length may be available to the neural decoder and could provide a shortcut that competes with semantic learning.
We estimate length bias by computing the Spearman rank correlation ($\rho$) between sentence length and the leading principal components of each embedding space.
Although this measure does not capture all forms of embedding sensitivity to sentence length, we find that it provides an effective empirical proxy for the extent to which sentence length is reflected in the global geometry of the embedding space.
A visual illustration of this analysis is provided in Appendix~\ref{sec:appendix_semantic_embedding_analysis}.
\paragraph{Intrinsic dimensionality.}
The target space should be learnable from limited MEG data.
We therefore prefer embedding spaces with lower intrinsic dimensionality, measured by effective rank, provided that they remain sufficiently expressive and reversible.

A further consideration is training-data transparency. In principle, a fully open embedding and inversion pipeline would be preferable, since it would allow us to verify that evaluation sentences were not present in the training data of either the embedding model or the inversion model. Among the embedding--inversion pairs we considered, however, we did not find a fully data-transparent option that also satisfied the practical requirements of expressivity and soft reversibility. We therefore control for possible corpus-leakage by comparing neural-based predictions against noise-control.
\begin{table*}[t]
\caption{
Comparison of candidate embedding spaces according to \textbf{expressivity}, \textbf{soft reversibility}, \textbf{length bias}, and \textbf{intrinsic dimensionality}.
ADA was chosen for its combination of high soft reversibility, low length bias, and good expressivity.
}
\label{tab:embedding_space_comparison}
\centering
\small
\setlength{\tabcolsep}{5pt}
\renewcommand{\arraystretch}{1.12}

\begin{tabular}{@{}lcccc@{}}
\toprule
\textbf{Embedding} 
& \textbf{Expressivity} $\uparrow$
& \textbf{Soft reversibility} $\uparrow$
& \textbf{Length bias} $\downarrow$
& \textbf{Intrinsic dim.} $\downarrow$ \\
\midrule

SONAR 
& $\mathbf{0.981 \pm 0.019}$
& $0.902$
& $0.819$
& $1024 / 197$ \\

T5 
& $0.813 \pm 0.018$ 
& $0.333$
& $0.569$ 
& $\mathbf{768 / 183}$ \\

\rowcolor{gray!10}
ADA 
& $0.888 \pm 0.031$ 
& $\mathbf{0.925}$
& $\mathbf{0.432}$ 
& $1536 / 195$ \\

\bottomrule
\end{tabular}
\end{table*}
\subsection{Objectives for Manifold Learning}
\label{subsec:objectives_semantic_manifold}

At the core of our training setup is a SigLIP-style contrastive loss \citep{zhai2023siglip, dAscoli2024DecodingWords}, which has been shown to be effective for aligning representations across modalities with different dimensionalities and statistical characteristics \citep{Radford2021CLIP}. However, in the low-data regime typical of non-invasive speech decoding, contrastive objectives alone are insufficient to learn the target manifold of semantic embeddings.

Previous brain-to-text and word-decoding approaches have largely relied on contrastive objectives to align neural signals with semantic embeddings \citep{Defossez2023DecodingSpeech,dAscoli2024DecodingWords}. However, we observe that in the low-data regimes typical of non-invasive speech decoding, contrastive loss primarily optimizes a retrieval objective. In this setting, the model learns a mapping that enables nearest-neighbor matching under cosine similarity \citep{minnema-herbelot-2019-brain}, but does not necessarily preserve the inter-vector distances or the scale of embedding magnitudes.

This limitation is problematic for our setting, where the goal is not merely to retrieve a correct target embedding, but to learn a mapping that faithfully captures the global geometry of the target semantic manifold. To address this, we draw inspiration from the manifold learning literature \citep{meila2023manifold} and introduce several auxiliary losses in addition to the SigLIP that force the model to learn the global properties of the target manifold and prevent collapse. The resulting training objective consists of the following components (invariance, covariance and variance losses are adopted from VICReg \citep{bardes2021vicreg}):

\textbf{SigLIP Loss}: 
A contrastive alignment term that formulates predicted--target matching as independent pairwise classification rather than a batch-wise softmax objective. This is useful for sentence-level semantic decoding, where different non-matching sentences may still be semantically related and should not necessarily be treated as mutually exclusive classes. We also found this objective more stable in the low-data MEG setting, particularly with small batches.
\begin{equation}
\mathcal{L}_{\mathrm{ctr}}
=
\frac{1}{n}
\sum_{i=1}^{n}
\sum_{j=1}^{n}
\log\!\left(
1 +
\exp\!\Big(
- t_{ij}
\big(
\tau\, \hat{\mathbf{y}}_i^{\top}\mathbf{y}_j + b
\big)
\Big)
\right),
\end{equation}
\textbf{Invariance Loss} : mean squared distance between predicted and target embeddings.

\begin{equation}
\mathcal{L}_{\mathrm{inv}}
=
\frac{1}{n}
\sum_{i=1}^{n}
\left\|
\mathbf{x}_i - \mathbf{y}_i
\right\|_2^2 .
\end{equation}

\textbf{Covariance Loss}: a decorrelation term that penalizes off-diagonal covariances between embedding dimensions, reducing redundancy and preventing informational collapse.
\begin{equation}
\mathcal{L}_{\mathrm{cov}}
=
\frac{1}{d}
\sum_{i \neq j}
C(\mathbf{x})_{i,j}^{2} .
\end{equation}

\begin{equation}
C(\mathbf{x})
=
\frac{1}{n-1}
\sum_{k=1}^{n}
(\mathbf{x}_k - \bar{\mathbf{x}})
(\mathbf{x}_k - \bar{\mathbf{x}})^{\top},
\end{equation}

\textbf{Variance Loss}: a hinge loss that enforces a minimum standard deviation across the batch for each embedding dimension, preventing collapse:
\begin{equation}
\mathcal{L}_{\mathrm{var}}
=
\frac{1}{d}
\sum_{j=1}^{d}
\max\!\left(0,\,
\gamma - \sigma_j(\mathbf{x})
\right) ,
\end{equation}

\begin{equation}
\sigma_j(\mathbf{x})
=
\sqrt{
\mathrm{Var}(x^{j}) + \epsilon
}.
\end{equation}

\textbf{Global Cosine Alignment Loss}: maximizes the average cosine similarity between predicted and target embedding vectors.
\begin{equation}
\mathcal{L}_{\mathrm{gcs}}
=
1
-
\frac{1}{n}
\sum_{i=1}^{n}
\frac{
\hat{\mathbf{y}}_i^{\top}\mathbf{y}_i
}{
\left\|\hat{\mathbf{y}}_i\right\|_2
\left\|\mathbf{y}_i\right\|_2
}.
\end{equation}

The final loss is:
\begin{equation}
\mathcal{L}
=
\alpha\,\mathcal{L}_{\mathrm{ctr}}
+
\beta\,\mathcal{L}_{\mathrm{gcs}}
+
\gamma\,\mathcal{L}_{\mathrm{inv}}
+
\delta\,\mathcal{L}_{\mathrm{var}}
+
\eta\,\mathcal{L}_{\mathrm{cov}} .
\end{equation}

To test whether each component contributes to the final decoding performance, we ablate individual terms from the training objective while keeping the rest of the pipeline fixed, as shown in Table~\ref{tab:loss_ablation_decoding_gain}.

\subsection{Inverting Semantic Embeddings Back to Text}
\label{subsec:inversion}

Embedding inversion \citep{Morris2023} is formulated as an iterative conditional generation problem, where the objective is to recover a text sequence $x^{*}$ given only its embedding $e^{*} = f(x^{*})$. The procedure initializes by sampling an initial hypothesis from a base generator,
\[
x_0 \sim p_\theta(x \mid e^{*}).
\]
At each iteration $t$, the current hypothesis $x_t$ is re-embedded to obtain $e_t = f(x_t)$, and a learned correction model generates an improved hypothesis conditioned on the current text and the embedding discrepancy:
\[
x_{t+1} \sim p_\phi(x \mid x_t, e_t, e^{*}).
\]
This iterative refinement progressively reduces the embedding distance $\lVert e_t - e^{*} \rVert$, yielding increasingly faithful reconstructions without direct optimization in discrete token space.
\subsection{Data}
For training and evaluation, we use LibriBrain \citep{ozdogan2025libribrain}, the largest single-subject speech-decoding MEG dataset available at the time of writing. Specifically, we use the Sherlock Holmes subset, which provides over 62 hours of MEG recordings from a single participant listening to continuous spoken narrative. The validation and test sets are held-out recording sessions, allowing us to evaluate generalization across sessions rather than across randomly sampled sentences.
\begin{table}[t]
\caption{
LibriBrain's Sherlock Holmes data splits used in our experiments. Validation and test sets are held-out recording sessions.
}
\label{tab:libribrain_stimuli}
\centering
\setlength{\tabcolsep}{4pt}

\begin{tabular}{lrrrr}
\toprule
\textbf{Stimuli} & \textbf{Words} & \textbf{Unique} & \textbf{Sentences} & \textbf{Hours} \\
\midrule
Sherlock Holmes Books (Train) & 600,107 & 20,837 & 40,659 & 61.80 \\
Sherlock Holmes Books (Validation) & 3,427 & 1,155 & 198 & 0.36 \\
Sherlock Holmes Books (Test) & 3,577 & 1,210 & 172 & 0.38 \\
Sherlock Holmes Books (Total) & 607,111 & 20,971 & 41,029 & 62.54 \\
\bottomrule
\end{tabular}
\end{table}

Text and audio were manually corrected, normalized, and force-aligned, with sentence boundaries defined by corpus punctuation. In this work, we prioritize dataset scale and semantic variability as key factors for semantic decoding, while deferring subject variability and cross-subject generalization to future studies. We chose MEG as our recording modality because it occupies a middle ground between fMRI and EEG: it offers high temporal resolution while providing substantially better spatial specificity than EEG. If MEG spatial resolution proves sufficient for capturing distributed semantic representations, this would open the possibility of jointly exploiting slow, distributed semantic signals and high-frequency auditory features within a single non-invasive modality.
\subsubsection{Preprocessing}
The recordings
were originally sampled at 1 kHz and downsampled to 250 Hz to preserve oscillations into the high-gamma range (70--125 Hz).
\section{Experiments}
We compare our results against prior \textit{Brain2Text} approaches using standard text-generation metrics: WER, BLEU, ROUGE, and BERTScore. 
While WER, BLEU, and ROUGE primarily measure lexical overlap and word-level reconstruction accuracy, BERTScore provides a complementary estimate of sentence-level semantic similarity. 
This is particularly important for our setting, where successful decoding may preserve the meaning of a sentence even when its exact wording is not recovered. At the same time, text-generation metrics alone cannot determine whether a decoded sentence was driven by neural information or by linguistic and dataset-level priors in the generation model. 
Following \citet{jo2024eegtotext}, we therefore include a noise-control analysis to estimate how much of the decoded output is attributable to the neural input rather than to textual priors alone.

Reversing semantic embeddings reliably requires learning a high-fidelity representation of the target semantic space; otherwise, inversion becomes infeasible. It is therefore critical to identify the minimum training-set size required for the method to become reliable. To this end, we report empirical scaling laws demonstrating that the fidelity of semantic-embedding mapping improves systematically with data scale, and we identify a minimum data regime beyond which the method becomes feasible.

\begin{table*}[t]
\caption{
Comparison of decoding methods (mean $\pm$ s.d.).
Brain2Semantics2Text and BrainECHO operate at the sentence level and do not rely on word-level alignment, whereas \citet{dAscoli2024DecodingWords} use word-aligned supervision.
Lexical-overlap metrics such as WER, BLEU-1, and ROUGE-1 therefore favor methods with access to word-level timing or lexical supervision, while BERTScore better reflects the sentence-level semantic reconstruction objective of our method.
}
\label{tab:methods_comparison}
\centering
\small
\setlength{\tabcolsep}{7pt}
\renewcommand{\arraystretch}{1.08}

\begin{tabular}{lcccc}
\toprule
\textbf{Method} 
& \textbf{WER} $\downarrow$ 
& \textbf{BLEU-1} $\uparrow$ 
& \textbf{ROUGE-1} $\uparrow$ 
& \textbf{BERTScore} $\uparrow$ \\
\midrule
\multicolumn{5}{l}{\textit{Sentence-level decoding, no word-level alignment}} \\
\addlinespace[2pt]
\rowcolor{gray!10}
Ours
& 1.925 $\pm$ 0.100 
& \textbf{0.100 $\pm$ 0.008} 
& \textbf{0.132 $\pm$ 0.003} 
& \textbf{0.830 $\pm$ 0.001} \\

Ours (noise control) 
& 2.648 $\pm$ 0.415 
& 0.086 $\pm$ 0.009 
& 0.113 $\pm$ 0.004 
& 0.819 $\pm$ 0.006 \\

BrainECHO 
& 1.018 $\pm$ 0.030 
& 0.061 $\pm$ 0.008 
& 0.091 $\pm$ 0.007 
& 0.828 $\pm$ 0.002 \\

BrainECHO (noise control) 
& 1.012 $\pm$ 0.005 
& 0.056 $\pm$ 0.006 
& 0.085 $\pm$ 0.008 
& 0.825 $\pm$ 0.002 \\

\midrule
\multicolumn{5}{l}{\textit{Word-level decoding with word-aligned supervision}} \\
\addlinespace[2pt]
d'Ascoli et al. 
& \textbf{0.871 $\pm$ 0.002} 
& \textbf{0.190 $\pm$ 0.004} 
& \textbf{0.172 $\pm$ 0.004} 
& 0.820 $\pm$ 0.001 \\

d'Ascoli et al. (noise control) 
& 0.994 $\pm$ 0.003 
& 0.072 $\pm$ 0.024 
& 0.057 $\pm$ 0.021 
& 0.794 $\pm$ 0.004 \\

\bottomrule
\end{tabular}
\end{table*}

\begin{figure}[t]
    \centering
    \includegraphics[
            width=1.0\columnwidth,
        trim=0cm 0cm 0cm 0cm,
        clip
    ]{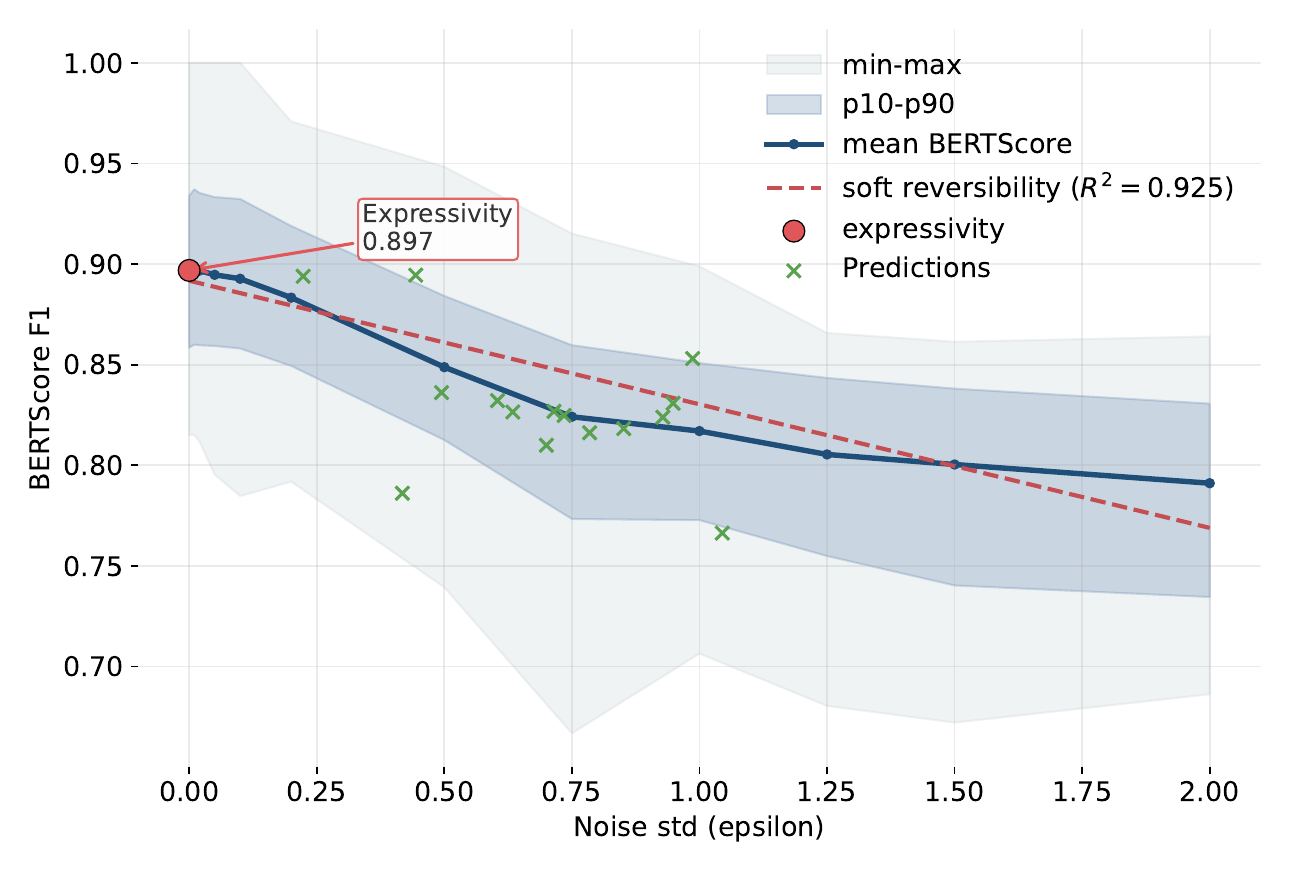}
\caption{
\textbf{Semantic embedding inversion.}
We evaluated semantic embedding inversion by adding controlled Gaussian noise with increasing standard deviation to the sentence embedding vectors and measuring the BERTScore F1 of the reconstructed text. 
Expressivity is defined as the reconstruction score obtained from the unperturbed embedding, corresponding to the ideal case of a perfectly predicted vector. 
Soft reversibility is measured by how smoothly reconstruction quality degrades as embedding noise increases, summarized by the \(R^2\) of the fitted relationship between noise level and BERTScore. 
Actual neural predictions are overlaid to show where model outputs fall along the resulting noise--performance curve (similar analyses for all candidate embeddings are provided in Appendix ~\ref{sec:appendix_semantic_embedding_analysis}).
}
    \label{fig:embedding_inversion}
\end{figure}

\begin{table}[t]
\caption{
Loss ablation results measured as mean decoded-sentence BERTScore F1 on the test set.
Values are mean $\pm$ standard deviation across runs.
Higher values indicate better decoded-sentence semantic similarity.
}
\label{tab:loss_ablation_decoding_gain}
\centering
\small

\begin{tabular}{lc}
\toprule
Method & BERTScore \\
\midrule
\textbf{Full objective (SigLIP + VICReg + global cosine)}
& $\mathbf{0.8297 \pm 0.0008}$ \\
\hspace{1em}w/o global cosine loss
& $0.8103 \pm 0.0077$ \\
\hspace{1em}w/o VICReg invariance term
& $0.8263 \pm 0.0025$ \\
\hspace{1em}w/o VICReg variance term
& $0.8267 \pm 0.0015$ \\
\hspace{1em}w/o VICReg covariance term
& $0.8270 \pm 0.0012$ \\
\bottomrule
\end{tabular}
\end{table}

\subsection{Results}
\begin{figure}[t]
    \centering
    \includegraphics[
            width=1.0\columnwidth,
        trim=0cm 0cm 0cm 0cm,
        clip
    ]{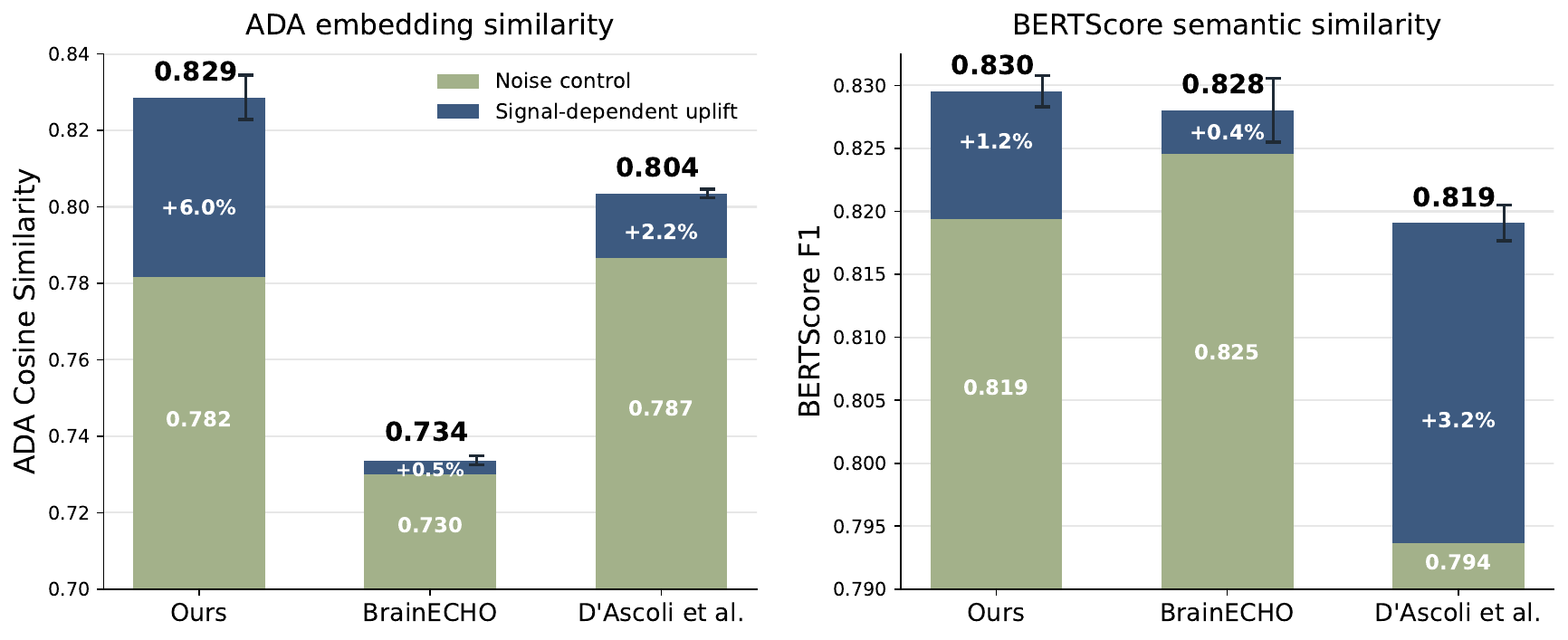}
    \caption{
        \textbf{Signal uplift.}
        Comparison of signal-dependent improvement over noise-control baselines across decoding methods.
    }
    \label{fig:signal_uplift_comparison}
\end{figure}

\begin{figure}[t]
    \centering
    \includegraphics[
            width=0.8\columnwidth,
        trim=0cm 0cm 0cm 0cm,
        clip
    ]{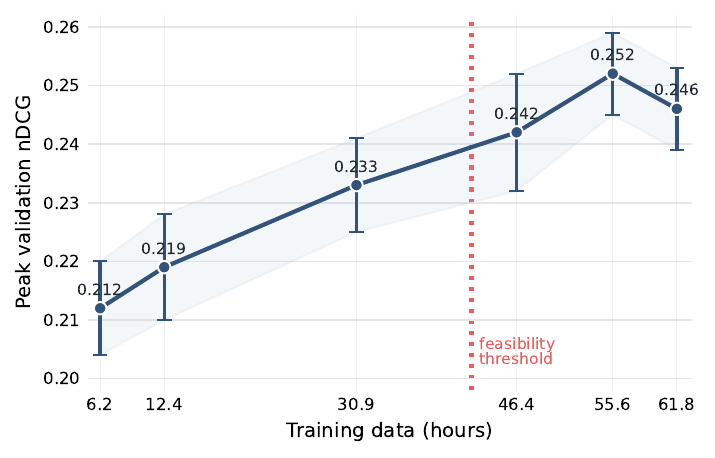}
    \caption{
        \textbf{Scaling Laws}
    }
    \label{fig:scaling_laws}
\end{figure}

\paragraph{Comparison of method performance.}
Table~\ref{tab:methods_comparison} compares the proposed \textit{Brain2Semantics2Text} approach with prior \textit{Brain2Text} decoding methods. The word-level metrics reveal a clear performance gap between word-level decoding, represented by d'Ascoli \citep{dAscoli2024DecodingWords}, and sentence-level decoding. This gap is expected, since word-level decoding operates in a much more constrained setting, with a closed vocabulary and access to exact word-level alignment. By contrast, when compared with acoustic-based sentence-level decoding, represented by BrainECHO \citep{Li2024BrainECHO}, our method performs favorably, achieving better results on BLEU-1, ROUGE-1, and BERTScore. These improvements hold both in absolute terms and when measured as neural-signal uplift relative to the noise control (Figure~\ref{fig:signal_uplift_comparison}). Our method tends to generate more words than appear in the ground-truth sentence, making WER less informative; recall-based metrics such as ROUGE-1, however, show a distinct signal-based improvement.
\paragraph{Signal uplift.}
Figure~\ref{fig:signal_uplift_comparison} evaluates the contribution of the neural signal across methods, measured using both BERTScore and ADA cosine similarity. BERTScore serves as the standard metric for comparing sentence-level reconstructions, while ADA cosine similarity provides a complementary measure of semantic similarity that more directly reflects the explicit training target of \textit{Brain2Semantics2Text}. Because all methods may exploit text priors and corpus-level regularities that are not driven by the brain signal, we measure performance relative to a noise-control baseline. This signal-dependent uplift estimates the contribution of the neural signal itself, rather than reconstruction quality attributable to the inversion model, language prior, or corpus-level bias.

On BERTScore, our method yields a 1.2-point uplift over its noise-control baseline, exceeding the uplift of the previous sentence-level method, BrainECHO, but remaining below the 3.2-point uplift of d'Ascoli et al. However, d'Ascoli et al. operate at the word level and require exact word-aligned supervision, whereas our method performs sentence-level decoding without word-level alignment. On ADA cosine similarity, our method shows the largest signal-dependent uplift, improving by 6.0 points and exceeding the corresponding uplift of the word-level method. This discrepancy between semantic metrics suggests that current evaluation measures capture different aspects of sentence-level decoding performance. More broadly, it highlights the need for better standardized metrics for evaluating semantic reconstruction in brain-decoding models.

\textbf{Scaling behavior of the Brain2Semantics2Text method.}
The method shows promising scaling behavior with increasing amounts of training data. Performance generally improved as training data increased, with gains appearing to saturate around 55.6 hours. These results indicate that the method makes effective use of additional MEG recordings of spoken language, supporting the potential value of larger-scale data collection while suggesting that future improvements may also benefit from increased data diversity.

We report performance using a retrieval-based metric: ``Discounted Cumulative Gain''~\citep{jarvelin2002cumulated}, rather than the text-based metrics, as below a certain performance threshold the full semantic inversion pipeline does not operate reliably and sentences reconstructions are not available. The retrieval metric evaluates the model’s ability to identify the closest matching sentence in the embedding space and, as such, does not capture global structural properties of the semantic manifold. Nevertheless, when all other factors remain constant, the retrieval performance provides a meaningful proxy for the overall effectiveness of the proposed method.
\section{Limitations and Future Work}
The current work represents an initial attempt at speech decoding through a direct mapping of MEG signals into semantic representations. Learning this semantic manifold proved to be challenging. Beyond the usual constraints of non-invasive neural recordings, such as low SNR and limited training data, reliable semantic encoding is likely to require greater semantic variability in the training corpus. We observed that the model learned semantic structure that was strongly biased toward the specific corpus used in this study. Therefore, future work should involve data collection protocols that emphasize variability in topics and concepts, perhaps guided by the properties of the target semantic manifold \citep{Pereira2018UniversalDecoder}.

The inversion methods used in this work were treated as black-box components and may not be optimally suited for semantic decoding. Further work is needed to better understand the ``soft reversibility'' of predicted vectors, introduced in the soft-reversibility analysis (Section~\ref{par:soft_reversibility}), and to optimize the preservation of semantic content, potentially by training inversion mechanisms specifically for this purpose.

The current work also does not address subject variability. Although this issue is beyond the present study, semantic representations may offer new routes for cross-subject generalization by modeling both shared semantic structure and participant-specific profiles in semantic representation space.

\section{Conclusion}
While fully non-invasive speech decoding remains a long-term goal, decoding contextual semantic content from brain activity represents an important step toward its realization. Neuroscientific evidence suggests that speech is processed across multiple levels of representation, from fast acoustic and lexical features to slower, higher-level semantic information. These levels may provide complementary targets for neural decoding. Recovering contextual semantics from MEG alongside lower-level speech information could therefore provide multiple sources of information for reconstruction, helping compensate for the limited signal quality of non-invasive neural recordings.\bibliography{paper}
\bibliographystyle{plainnat}



\clearpage
\appendix
\onecolumn

\section{Impact Statement}
\label{sec:appendix_impact_statement}

This paper presents work toward non-invasive speech decoding, with potential applications in brain-computer interfaces for individuals who have lost the ability to speak. Clinical deployment remains distant, as current performance is still below what communication aids require, and substantial further work is needed. We also note that neural decoding technologies raise clear privacy concerns, since they involve inferring mental content from brain activity. Our work uses only publicly available research datasets with their own ethics approvals and decodes perceived speech rather than covert thought. As decoding capabilities improve, the field will need norms around consent, data ownership, and the boundary between assistive and surveillant applications. These are questions we do not resolve here, but consider essential.

\section{Hyperparameters}
\label{sec:appendix_hyperparameters}

\vspace{0.75em}
\begin{center}
\captionof{table}{Hyperparameters used for MEG-to-semantic embedding training.}
\label{tab:meg2sem_hparams}
\begin{tabular}{ll}
\toprule
\textbf{Parameter} & \textbf{Value} \\
\midrule
Dropout & 0.6 \\
Learning rate & $3 \times 10^{-5}$ \\
Optimizer & AdamW \\
Loss type & SigLIP \\
VICReg weight & 5.0 \\
Cosine loss weight & 6.0 \\
Contrastive loss weight & 2.0 \\
Aggregation & Attention, 4 heads \\
\bottomrule
\end{tabular}
\end{center}
\vspace{-0.7em}
\section{Compute Resources}
\label{sec:compute_resources}

\vspace{-0.5em}
All MEG-to-semantic embedding models were trained on a single NVIDIA GPU using 4 CPU cores and 64 GiB of system memory. A typical full-data run took approximately 16--18 GPU-hours, corresponding to about 6 minutes per epoch for 150--170 epochs. Final training across seven random seeds used approximately 110--130 GPU-hours, excluding exploratory runs, failed jobs, and downstream evaluation or decoding.
\section{Decoding Examples}
\label{sec:appendix_qualitative_examples}

\begin{figure}[H]
    \centering
    \rotatebox{-90}{%
        \includegraphics[
            height=0.62\textheight,
            trim=4.2cm 0cm 4.5cm 0cm,
            clip
        ]{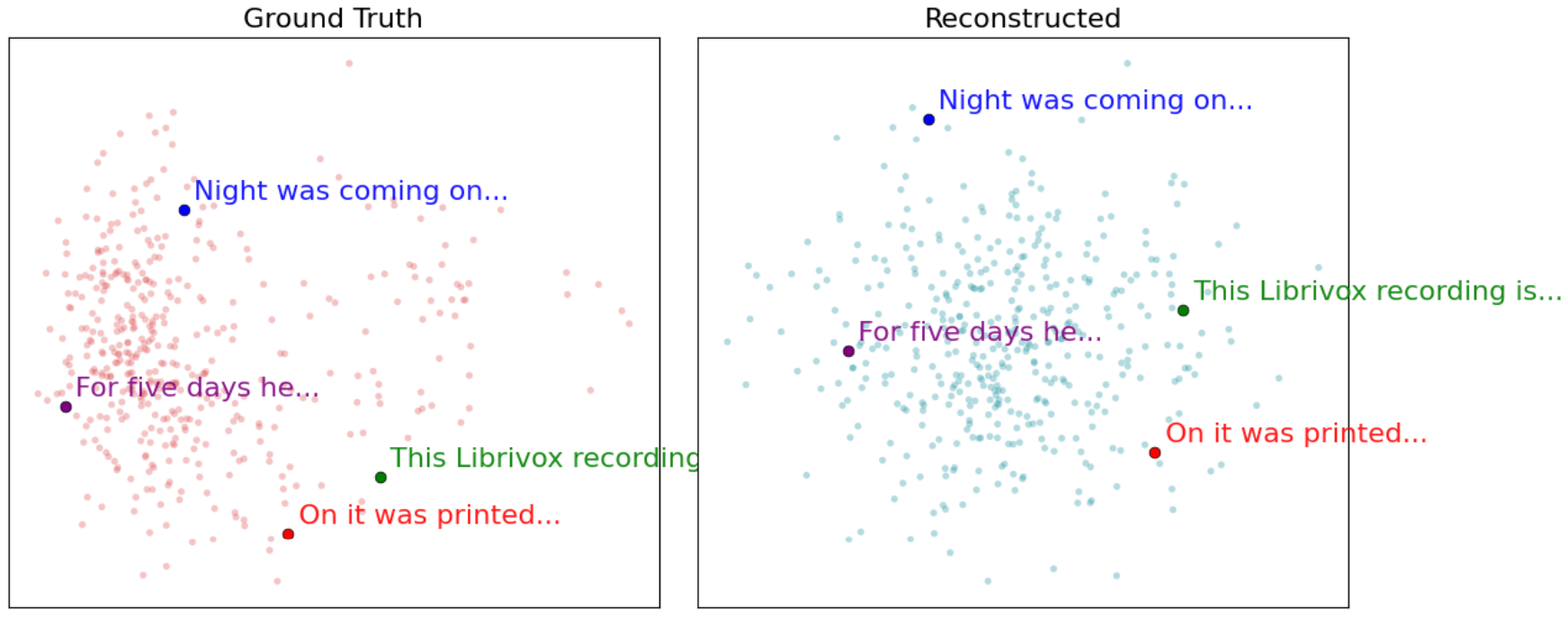}
    }
    \caption{
        \textbf{Semantic geometry of reconstructed sentence embeddings.}
        PCA projections of sentence-level embeddings for ground-truth sentences and reconstructed predictions. Highlighted examples illustrate that relative positions and global structure are largely preserved under reconstruction. Examples are shown for geometric comparison only, and textual reconstructions may differ from the ground truth.
    }
    \label{fig:semantic_geometry}
\end{figure}

\begin{figure}[H]
    \centering
    \includegraphics[width=0.97\linewidth]{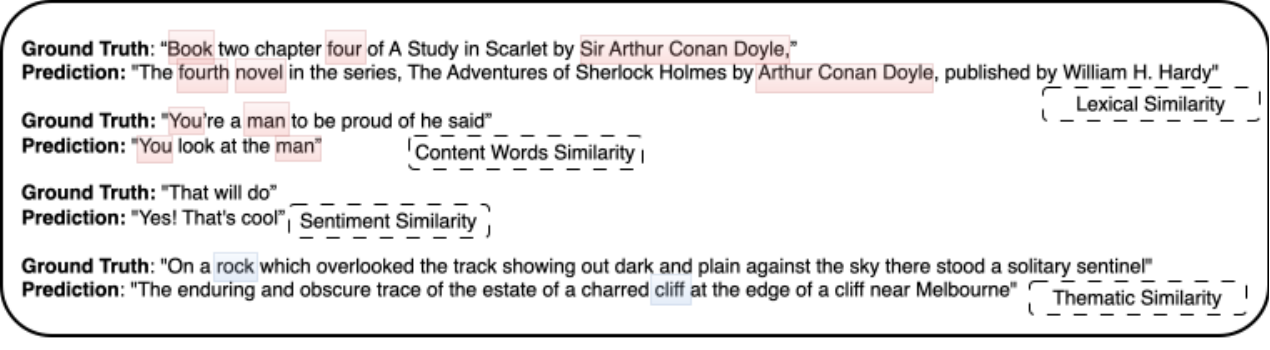}
    \caption{
        \textbf{Qualitative decoding examples.}
        Example decoded sentences shown for qualitative illustration of semantic similarity between the target sentence and the reconstructed output.
    }
    \label{fig:qualitative_decoding_examples}
\end{figure}
\section{Neural Signal Ablations}
\label{sec:appendix_neural_signal_ablations}

\vspace{-1.2em}

\begin{figure}[H]
    \centering
    \rotatebox{270}{%
        \includegraphics[
            width=0.2\textheight,
            trim=7cm 0cm 4cm 0cm,
            clip
        ]{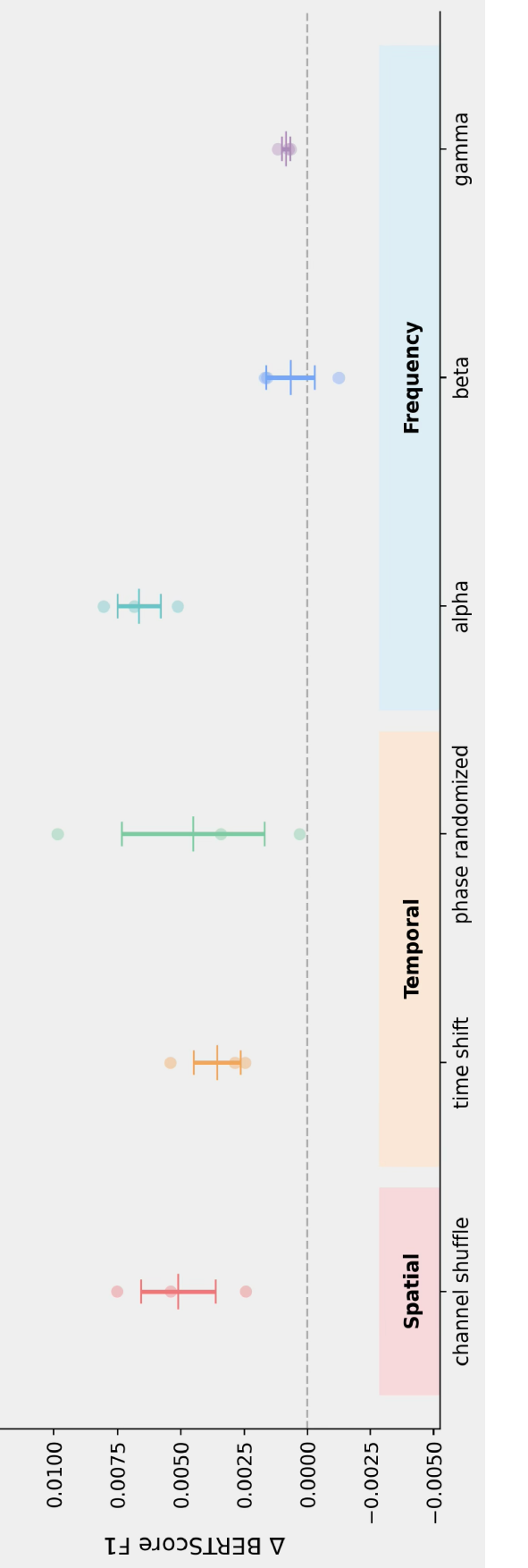}
    }
    \vspace{-1.2em}
    \caption{
        \textbf{Neural signal properties supporting semantic decoding.}
        We examine how different properties of the MEG signal affect semantic mapping performance, providing an interpretable view of which aspects of the neural response contribute most to decoding.
    }
    \label{fig:signal_ablations}
\end{figure}

\vspace{-0.5em}

\section{Semantic Embedding Analysis}
\label{sec:appendix_semantic_embedding_analysis}

\vspace{-1.2em}

\begin{figure}[H]
    \centering
    \includegraphics[
        width=1.0\linewidth,
        trim=0cm 0.3cm 0cm 0.7cm,
        clip
    ]{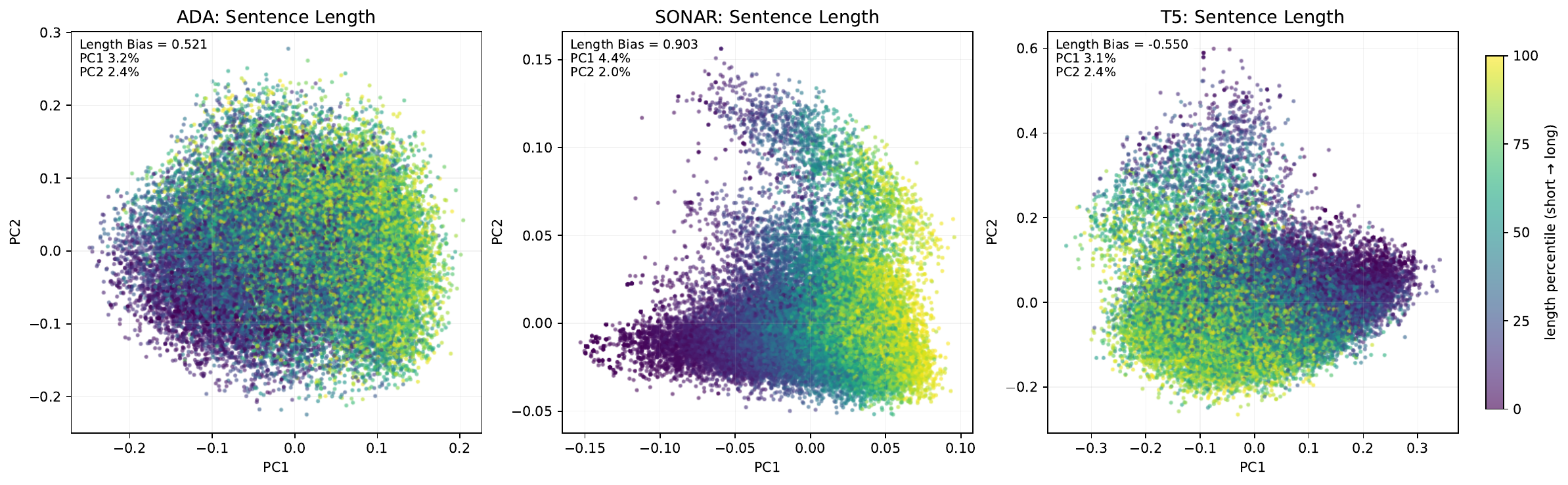}
    \vspace{-0.8em}
    \caption{
        \textbf{PCA comparison of ADA and SONAR semantic embeddings, colored by sentence length.}
        While SONAR embeddings segregate sentences by length, forming length-dependent regions in the embedding space, ADA embeddings remain more invariant to sentence length. This suggests that ADA is less vulnerable to the sentence-length confound.
    }
    \label{fig:pca_embeddings_length}
\end{figure}

\vspace{-0.5em}

\begin{figure}[H]
    \centering
    \includegraphics[
        width=1.0\linewidth,
        trim=0cm 0.3cm 0cm 0.7cm,
        clip
    ]{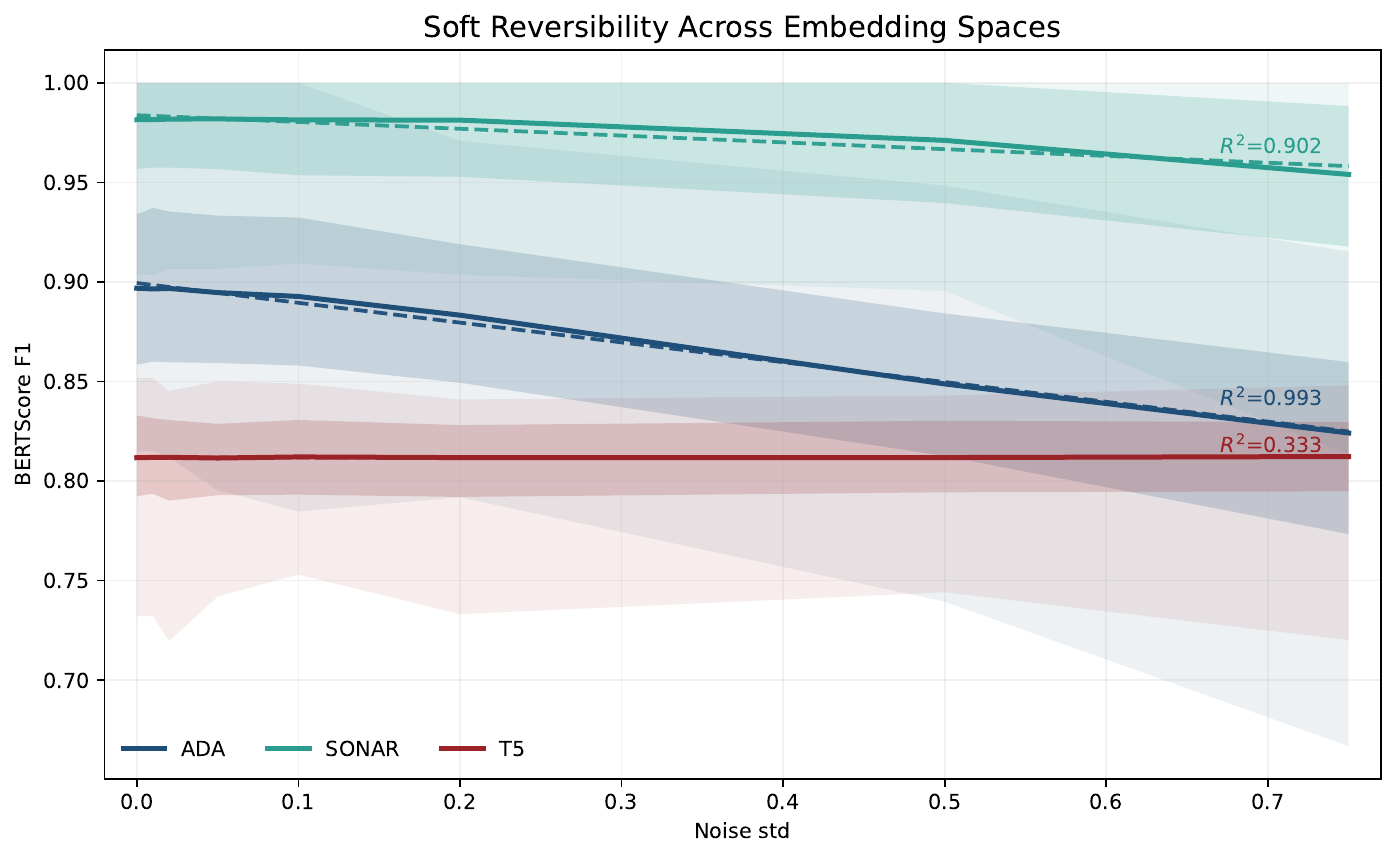}
    \vspace{-0.8em}
    \caption{
        \textbf{Comparison of candidate embedding spaces across embedding-space diagnostics.}
        Candidate semantic embedding spaces are compared according to expressivity, soft reversibility, length bias, and intrinsic dimensionality. A full explanation of how these measures are computed is provided in the appendix.
    }
    \label{fig:embedding_space_diagnostics}
\end{figure}

\end{document}